\pdfoutput=1

\documentclass[11pt]{article}

\usepackage{microtype}

\usepackage[final]{acl}

\usepackage{times}
\usepackage{latexsym}

\usepackage[T1]{fontenc}

\usepackage[utf8]{inputenc}

\usepackage{microtype}

\usepackage{inconsolata}

\usepackage{graphicx}

\usepackage{inconsolata}

\usepackage{booktabs}
\usepackage{graphicx}
\usepackage{amsmath}

\usepackage{anyfontsize}

\newcommand{\langp}[2]{#1$\rightarrow$#2}
\newcommand{\langpb}[2]{#1$\leftrightarrow$#2}

\newcommand{\opus}{\textsc{OPUS}}
\newcommand{\nllb}{\textsc{NLLB}}

\newcommand{\ti}{\textsc{TI}}

\newcommand{\cgpt}{\textsc{GPT-3.5}}
\newcommand{\flores}{FLORES}

\newcommand{\dqe}{$\Delta$QE}
\newcommand{\mathqe}{\text{QE}}
\newcommand{\mathdqe}{\Delta\text{QE}}

\title{Did Translation Models Get More Robust Without Anyone Even Noticing?}

\author{Ben Peters\textsuperscript{$\ast$} \and
        Andr\'e F.~T. Martins\textsuperscript{$\ast\dag\ddag\diamond$} \\
\textsuperscript{$\ast$}Instituto de Telecomunica\c{c}\~oes, Lisbon, Portugal \\
\textsuperscript{$\dag$}Instituto Superior Técnico, Universidade de Lisboa, Lisbon, Portugal \\
\textsuperscript{$\ddag$}ELLIS Unit Lisbon (LUMLIS), Lisbon, Portugal\\
\textsuperscript{$\diamond$}Unbabel, Lisbon, Portugal\\
\href{mailto:benzurdopeters@gmail.com}{\tt benzurdopeters@gmail.com},\quad
\href{mailto:andre.t.martins@tecnico.ulisboa.pt}{\tt andre.t.martins@tecnico.ulisboa.pt}
}

\begin{document}
\maketitle
\begin{abstract}
Neural machine translation (MT) models achieve strong results across a variety of settings, but it is widely believed that they are highly sensitive to ``noisy'' inputs, such as spelling errors, abbreviations, and other formatting issues. 
In this paper, we revisit this insight in light of recent multilingual MT models and large language models (LLMs) applied to machine translation.
Somewhat surprisingly, we show through controlled experiments that these models are far more robust to many kinds of noise than previous models, even when they perform similarly on clean data.
This is notable because, even though LLMs have more parameters and more complex training processes than past models, none of the open ones we consider use any techniques specifically designed to encourage robustness.
Next, we show that similar trends hold for social media translation experiments -- LLMs are more robust to social media text.
We include an analysis of the circumstances in which source correction techniques can be used to mitigate the effects of noise.
Altogether, we show that robustness to many types of noise has increased.
\end{abstract}

\section{Introduction}
For years, the conventional wisdom has been that neural machine translation (MT) models are highly sensitive to source-side artificial and natural noise at inference time \citep{belinkovsynthetic}.
This insight has motivated many works that seek to make MT models more robust to noise through either specialized training \citep{ebrahimi-etal-2018-adversarial,karpukhin-etal-2019-training,park-etal-2020-adversarial,vaibhav-etal-2019-improving} or bespoke architectures \citep{rust2022language,salesky-etal-2021-robust}.
However, MT is increasingly being performed in a different paradigm than when these analyses and architectures were created.
Previously, models were mostly trained from scratch on task-specific data.
Nowadays, strong results often depend on
instruction-tuned large language models (LLMs) like TowerLLM \citep{alves2024tower} or opaque proprietary systems like ChatGPT.\footnote{\url{https://chat.openai.com/}}
These huge models may make existing robustness techniques more expensive (due to higher parameter counts) or impossible (specialized architectures cannot be grafted onto an existing pretrained system).
So the question is, are these robustness techniques still necessary in the era of LLMs, or have larger models and training sets made today's models sufficiently robust on their own?

\begin{figure}
    \centering
    \includegraphics[width=1.0\linewidth]{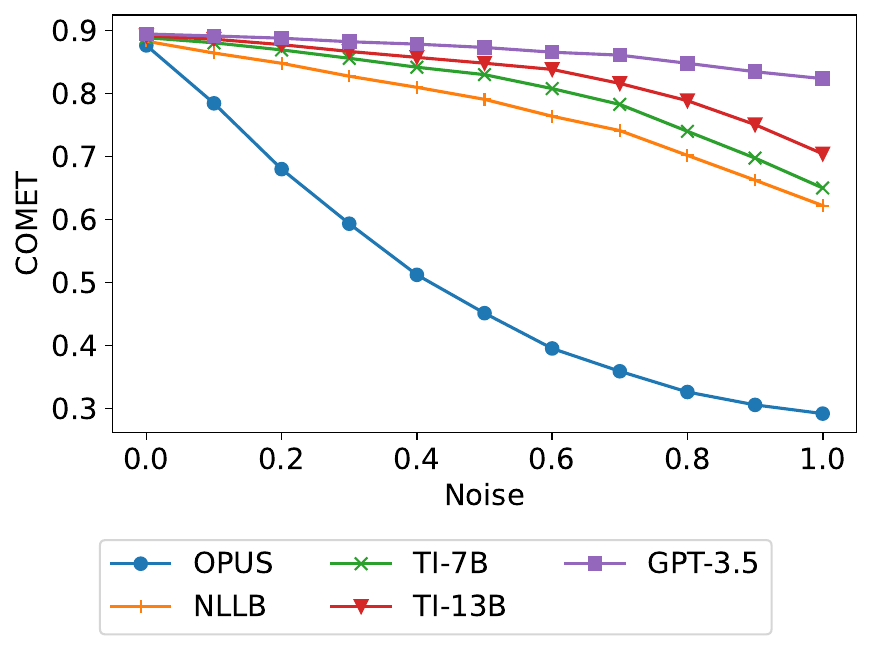}
    \caption{COMET-22 on the FLORES English-French devtest set when some proportion of source tokens are noised by swapping an adjacent pair of characters.}
    \label{fig:en-fr-swap}
\end{figure}

In this work, we investigate these questions through experiments on social media text and synthetically noised corpora.
These experiments play complementary roles: social media text contains diverse noise phenomena, but their effect is hard to isolate because the errors are unlabeled.
On the other hand, synthetic errors differ from real-world noise, but they are \textbf{interpretable} and \textbf{controllable}, offering a way to measure noise \emph{in vitro}.
By evaluating on a broad spectrum of error types, we can paint a more vivid picture of what kinds of noise, and at what quantities, cause problems for MT systems.
We make the following contributions:\footnote{Our code is available at \url{https://github.com/utter-project/robust-mt}.}

\begin{itemize}
    \item We show (\S \ref{section:synth-robustness}) that large pretrained models are much more robust to synthetic errors than conventional NMT models (see Figure~\ref{fig:en-fr-swap}), even when they perform similarly on clean data. This result holds across noise types and language pairs, even though the large models lack architectural features that specifically encourage robustness to character noise.
    \item We introduce (\S \ref{subsec:synth-exp}) a novel technique for measuring the robustness of MT models by learning a regression to predict the quality decline as a function of how noisy the source is.
    \item We show (\S \ref{subsec:mtnt-exp}) that models that are robust to synthetic errors perform better at translating social media text.
    We investigate the relationship between synthetic robustness and performance on ``real-world'' noise.
    \item We conduct (\S \ref{subsec:multilexnorm-exp}) reference-free MT experiments on MultiLexNorm \citep{van-der-goot-etal-2021-multilexnorm},
    which has never before been used for MT.
    We show that LLMs are more robust than conventional models to this type of noise.
    \item We show (\S \ref{section:correction}) that \textbf{finetuning on noisy translation data} and \textbf{source correction pipelines} are both effective approaches to mitigate synthetic noise without substantially worsening performance on clean data, allowing conventional NMT models to become more robust than GPT-3.5 to 3 out of 4 synthetic noise types.
    Combining correction with 7-13B parameter LLM-based translation models yields even higher robustness, allowing these pipelines to surpass GPT-3.5 on all of our synthetic benchmarks, often by a wide margin.
    Although correction is less effective for social media data on the whole, 
    many individual examples benefit from it, suggesting that identifying these examples is a future direction.
\end{itemize}

\section{Background}

\subsection{Architectures for MT}

\paragraph{The transformer.}
In recent years, mainstream MT techniques have been based on the transformer \citep{vaswani2017attention}, which uses multi-headed self-attention to mix information across time steps.
In the original work, transformers used an encoder-decoder paradigm similar to recurrent MT models \citep{bahdanau2014neural}.
These models pair an encoder over the source with a decoder, an autoregressive language model that predicts target tokens one at a time.
These tokens usually come from a subword vocabulary \citep{kudo-2018-subword,sennrich-etal-2016-neural}.
Initially, transformer MT models were trained from scratch for a single language pair on parallel data from sources such as the OPUS parallel corpus collection \citep{tiedemann-2012-parallel}.

\paragraph{Multilingual models.}
Although single language pair models often perform well, they struggle in the absence of large quantities of data, making it difficult to achieve good results in low resource language settings.
This problem can be mitigated through multilingual training with systems like M2M-100 \citep{fan2021beyond} and NLLB-200 \citep{nllbteam2022language}.
Low resource language pairs often benefit from training data in other languages.
One challenge is language imbalance -- the subword vocabulary and training procedure need to be designed to allow strong performance across covered language pairs in spite of this imbalance.

\paragraph{LLMs for MT.}
In parallel to these MT-centric developments, transformers have increasingly been used in a transfer learning set-up in which a model is pretrained on a generic objective for which massive data is available.
The model can then be finetuned on one or more downstream tasks.
When the pretraining objective is language modeling \citep{radfordimproving}, it is straightforward to use the model for generation tasks, including MT \citep{hendy2023gptmt}.
Recently, the paradigm has shifted from traditional finetuning to instruction tuning \citep{sanh2022t0,wei2022flan}, in which the finetuning data is accompanied by an instructional prompt.
This has been shown to give models the ability to generalize to related tasks and has proven effective for MT \citep{alves-etal-2023-steering,alves2024tower}.

\subsection{Robustness to Character Noise}
Character perturbations can have a large negative impact on MT model performance \citep{belinkovsynthetic}.
Consequently, a number of techniques have been proposed to mitigate their impact.

\paragraph{Robustness through training.}
A common technique to increase robustness is to train MT models on examples with added source errors.
Given that high-quality corpora containing authentic errors are rare,
the added noise is generally synthetic \citep{karpukhin-etal-2019-training}, although it can be tuned to resemble natural errors \citep{martucci2021lexical,vaibhav-etal-2019-improving}.
Whether training on synthetic noise is actually helpful for becoming robust to natural errors is an open question, with various works coming to contradictory conclusions \citep{belinkovsynthetic,vaibhav-etal-2019-improving}.

\paragraph{Robustness through architecture.}
As an alternative to specialized training techniques, robustness can be achieved with architectures other than the ubiquitous subword-level transformer.
Modeling at the character or byte level \citep{sutskever2011generating,xue-etal-2022-byt5} means that perturbations make only small changes to the sequence of tokens that the model is exposed to, whereas these same perturbations can cause a subword-level model to be exposed to completely different subword types.
This may make character- and byte-level models more robust, although the evidence is mixed \citep{mielke2021between}.
These models are also much slower than subword-level models because of longer sequence lengths.
As an alternative, MT models can be trained on representations that are invariant to character shuffles \citep{belinkovsynthetic} or on visual representations of text \citep{salesky-etal-2021-robust}.
\section{Robustness to Synthetic Noise}
\label{section:synth-robustness}

\begin{table}[t]
    \centering
    \small
\textbf{\langp{xx}{en}}\\
\begin{tabular}{lrrrrr}
\toprule
Model & \langp{de}{en} & \langp{fr}{en} & \langp{ko}{en} & \langp{pt}{en} & avg.\\
\midrule
\opus    & 88.17 & 89.19 & 86.35 & 88.36 & 88.02\\
\nllb    & 89.28 & 89.29 & 87.69 & 89.72 & 89.00\\
\ti   & \textbf{89.77} & \textbf{89.76} & \textbf{88.69} & \textbf{90.16} & \textbf{89.60}\\
\cgpt & 89.64 & 89.45 & 87.98 & 89.81 & 89.22\\
\bottomrule
\end{tabular}

\bigskip

\textbf{\langp{en}{xx}}\\
\begin{tabular}{lrrrrr}
\toprule
Model & \langp{en}{de} &  \langp{en}{fr} &  \langp{en}{ko} & \langp{en}{pt} & avg. \\
\midrule
\opus    & 84.02 &  87.63 & 86.58 &  88.94 & 86.79\\
\nllb    & 88.07 & 88.30 & 88.48 & 89.58 & 88.61\\
\ti & \textbf{88.57} &  \textbf{89.16} &  \textbf{90.12} &  \textbf{90.02} & \textbf{89.47}\\
\cgpt & 88.52 & 88.83 & 89.04 & 89.83 & 89.05\\
\bottomrule
\end{tabular}
    \caption{COMET on \flores{} without added noise.}
    \label{table:clean-flores}
\end{table}

In our first experiments, we evaluate how models perform in the presence of token-level synthetic errors.
Although these errors differ from ``naturally occurring'' noise, they are adjustable and function as a stress test for MT systems.

\subsection{Experiments}
\label{subsec:synth-exp}
In all of our synthetic experiments, we adopt a simple set-up: for each translation corpus, we introduce a particular type of perturbation into some percentage of the source-side tokens.
We then compare performance translating this perturbed corpus to the performance on clean data.
A model's robustness can be characterized by the steepness of its decline as the noise level is increased: a flatter slope indicates that the model handles noise better.

\paragraph{Data.}
We use four types of synthetic perturbations, each of which is a plausible error based on the mechanics of typing.
For each noise type, we create ten noised versions of the FLORES-200 devtest set \citep{nllbteam2022language} corresponding to noise levels $p \in \{0.1, 0.2, \dots 1.0\}$.
Within a version of the corpus, each whitespace-delimited token is perturbed with probability $p$ and otherwise not altered.
Therefore each token can be perturbed at most once.
We use the following noise types:

\begin{itemize}
    \item \textbf{swap}: flip two adjacent characters.
    \item \textbf{dupe}: duplicate a character.
    \item \textbf{drop}: delete a character.
    \item \textbf{key}: replace a character with an adjacent key. Further details are in Appendix~\ref{appendix:key-noise}.
\end{itemize}

\paragraph{Models.}
We use models that differ in their scope (bi- or multilingual), architecture (encoder-decoder or decoder-only), and size (74M-13B parameters).

\begin{itemize}
    \item \opus: We use transformer encoder-decoder models trained from scratch on a single language pair and released as part of OPUS-MT \citep{tiedemann-thottingal-2020-opus}. Model and vocabulary sizes are listed in Appendix~\ref{appendix:opus-model}.
    \item \nllb{} \citep{nllbteam2022language}, like \opus{}, is an encoder-decoder transformer trained on parallel text.
    However, \nllb{} is a many-to-many system trained on data in 202 languages.
    We use the 3.3 billion parameter version.
    \item \textsc{TI}: We use the 13 billion parameter version of TowerInstruct-v0.1 \citep{alves2024tower}, an instruction-tuned LLM that can translate between 10 languages.
    \item \cgpt:\footnote{Specifically, we use \texttt{gpt-3.5-turbo-1106}.} the architecture and training data of \cgpt{} are unknown, but we include it because of its success at MT \citep{hendy2023gptmt}; the related \textsc{GPT-4} can also correct character perturbations \citep{cao-etal-2023-unnatural}.
\end{itemize}

\noindent As \ti{} and \cgpt{} were both trained on closed data, it is possible that they were trained on our test sets.
We include them because of the lack of high-quality fully-open LLM-based translation systems.

\begin{table}[t]
    \small
    \centering

\textbf{swap}\\
\begin{tabular}{lrrrrr}
\toprule
Model & \langp{de}{en} & \langp{fr}{en} & \langp{ko}{en} & \langp{pt}{en} & avg.\\
\midrule
\opus  &  -65.05 & -65.41  & -36.18  & -63.44 & -57.52\\
\nllb  &  -18.14 & -20.97  & -23.79  & -20.81 & -20.93\\
\ti &  -27.61 & -27.01  & -23.45  & -25.54 & -25.90\\
\cgpt  &   \textbf{-4.36}  &  \textbf{-5.85} & \textbf{-20.89} &  \textbf{-6.78} & \textbf{-9.47}\\
\bottomrule
\end{tabular}

\textbf{drop}\\
\begin{tabular}{lrrrrr}
\toprule
Model & \langp{de}{en} & \langp{fr}{en} & \langp{ko}{en} & \langp{pt}{en} & avg.\\
\midrule
\opus &  -53.61 & -49.92  & -29.54 & -52.48 & -46.39\\
\nllb  & -16.50  & -17.27  & -21.11  & -19.04 & -18.48\\
\ti & -19.62  & -18.01  & \textbf{-17.33} &  -17.71 & -18.17\\
\cgpt &  \textbf{-6.55}   & \textbf{-5.68}  & -17.81 & \textbf{-7.09} & \textbf{-9.28}\\
\bottomrule
\end{tabular}

\textbf{dupe}\\
\begin{tabular}{lrrrrr}
\toprule
Model & \langp{de}{en} & \langp{fr}{en} & \langp{ko}{en} & \langp{pt}{en} & avg.\\
\midrule
\opus &  -34.31 &  -31.83 &  -6.92 &  -34.58 & -26.91\\
\nllb &  -4.07 &  -5.31 &  -4.36 &  -4.58 & -4.58\\
\ti &  -3.37 &  -4.54 &  \textbf{-1.82} &  -3.62 & -3.38\\
\cgpt &   \textbf{-1.36}    &  \textbf{-1.42}   & -5.64  & \textbf{-1.44} & \textbf{-2.47}\\
\bottomrule
\end{tabular}

\textbf{key}\\
\begin{tabular}{lrrrrr}
\toprule
Model & \langp{de}{en} & \langp{fr}{en} & \langp{ko}{en} & \langp{pt}{en} & avg.\\
\midrule
\opus  & -63.78  & -65.40 & -38.48  & -65.50 & -58.29\\
\nllb  & -20.66 &  -21.86  & -28.01  & -23.60 & -23.53\\
\ti &  -29.15 &  -32.18  & -19.80  & -34.15 & -28.82\\
\cgpt &    \textbf{-9.17}   &  \textbf{-8.63} & \textbf{-16.31} & \textbf{-10.27} & \textbf{-11.09}\\
\bottomrule
\end{tabular}

    \caption{COMET-slope on \flores{} for \langp{xx}{en}.}
    \label{table:flores-slope-xx-en}
\end{table}

\begin{table}[t]
    \small
    \centering

\textbf{swap}\\
\begin{tabular}{lrrrrr}
\toprule
Model & \langp{en}{de} & \langp{en}{fr} & \langp{en}{ko} & \langp{en}{pt} & avg.\\
\midrule
\opus & -72.01  &  -69.59 &  -73.99 &  -72.97 & -72.14\\
\nllb & -23.33  &  -22.75 &  -19.68 &  -22.41 & -22.04\\
\ti & -16.64 &  -13.71 &  -12.63 &  -13.44 & -14.11\\
\cgpt &  \textbf{-3.89} &  \textbf{-4.46} &  \textbf{-4.79} &  \textbf{-3.76} & \textbf{-4.23}\\
\bottomrule
\end{tabular}

\textbf{drop}\\
\begin{tabular}{lrrrrr}
\toprule
Model & \langp{en}{de} & \langp{en}{fr} & \langp{en}{ko} & \langp{en}{pt} & avg.\\
\midrule
\opus & -67.77 &  -63.30  &  -71.31 &  -69.66 & -68.01\\
\nllb & -22.65 &  -22.23  &  -18.45 &  -21.71 & -21.26\\
\ti & -17.22 &  -15.00 &  -9.08 &   -14.68 & -14.00\\
\cgpt & \textbf{-6.59}  &  \textbf{-7.32}  & \textbf{-6.72}  & \textbf{-6.63} & \textbf{-6.81}\\
\bottomrule
\end{tabular}

\textbf{dupe}\\
\begin{tabular}{lrrrrr}
\toprule
Model & \langp{en}{de} & \langp{en}{fr} & \langp{en}{ko} & \langp{en}{pt} & avg.\\
\midrule
\opus  & -54.90 & -46.25 & -65.86 & -57.94 & -56.24\\
\nllb  & -4.04 & -3.81  & -2.79 & -4.13 & -3.69\\
\ti & -2.40 & -1.79 &  \textbf{-1.39} & -1.89 & -1.87\\
\cgpt & \textbf{-1.14}  & \textbf{-1.32} & -1.42  &  \textbf{-0.98} & \textbf{-1.21}\\
\bottomrule
\end{tabular}

\textbf{key}\\
\begin{tabular}{lrrrrr}
\toprule
Model & \langp{en}{de} & \langp{en}{fr} & \langp{en}{ko} & \langp{en}{pt} & avg.\\
\midrule
\opus & -72.46 & -72.01 & -76.64 & -75.81 & -74.23\\
\nllb & -27.32 &  -25.91 &  -23.90 &  -25.57 & -25.67\\
\ti & -24.51 &  -21.10 & -15.95 &  -22.04 & -20.90\\
\cgpt & \textbf{-8.19}  &  \textbf{-8.17} &  \textbf{-8.91} & \textbf{-7.78} & \textbf{-8.26}\\
\bottomrule
\end{tabular}

    \caption{COMET-slope on \flores{} for \langp{en}{xx}.}
    \label{table:flores-slope-en-xx}
\end{table}

\paragraph{Inference.}
For \cgpt, we sample with temperature 0.
For other models, we decode with a beam size of 5.
Details are in Appendix~\ref{appendix:inference}.

\paragraph{Evaluation.}
Our base corpus-level translation metric is COMET \citep{rei-etal-2020-comet}.\footnote{We use \texttt{Unbabel/wmt22-comet-da} \citep{rei-etal-2022-comet}.}
COMET computes a normalized score for a hypothesis $y$, conditioned on the source $x$ and a reference $r$.
When we compute scores for translations from noisy data, we provide the COMET model the clean source, not the noisy version that was actually used to generate hypotheses.
We measure the \textbf{trajectory} of performance as the amount of noise is increased, as depicted in Figure~\ref{fig:en-fr-swap}.
To represent this trajectory as a single number, we fit a linear regression to predict the COMET decline relative to the clean score\footnote{There is no need to learn an intercept term because the decline is relative to the model's clean performance.} as a function of the proportion of noised tokens.
We report the learned slope, which we call COMET-slope.
A higher (closer to zero) COMET-slope indicates a more robust model.
This metric can be interpreted as the number of COMET points that would be lost if every token were corrupted.

\paragraph{Results.}
Table~\ref{table:clean-flores} shows that on clean data, \ti{} records the highest COMET for all eight language pairs.
The gap between the strongest system and the much smaller \opus{} models is less than $2.5$ COMET for all pairs except \langp{en}{de} and \langp{en}{ko}.
However, Tables~\ref{table:flores-slope-xx-en} and \ref{table:flores-slope-en-xx} show that \opus{} suffers more from perturbations than the other models do.
On the other end of the spectrum, \cgpt{} is almost always the most robust system.
\nllb{} and \ti{} are between these two extremes.
For swap and drop noise, \nllb{} is more robust than \ti{} when translating to English, while the reverse is true when translating from English.
This trend is less consistent for dupe noise.
For key noise, \nllb{} is more robust than \ti{} for every pair except \langp{ko}{en}.
BLEU and chrF results are in Appendix~\ref{appendix:synth-metrics}.

\subsection{Analysis}

\begin{table}[t]
    \small
    \centering
    \begin{tabular}{lrrrrrr}
    \toprule
        Model    & clean & swap & drop & dupe & key\\
        \midrule
        \opus   & 88.94 & -72.97 & -69.66 & -57.94 & -75.81\\
        OPUSLLM & 85.78 & -73.05 & -69.68 & -55.63 & -74.48\\
        \bottomrule
    \end{tabular}
    \caption{Robustness of OPUSLLM on \langp{en}{pt}.}
    \label{table:opusllm}
\end{table}

\begin{figure}[t]
    \centering
    \includegraphics[width=1.0\linewidth]{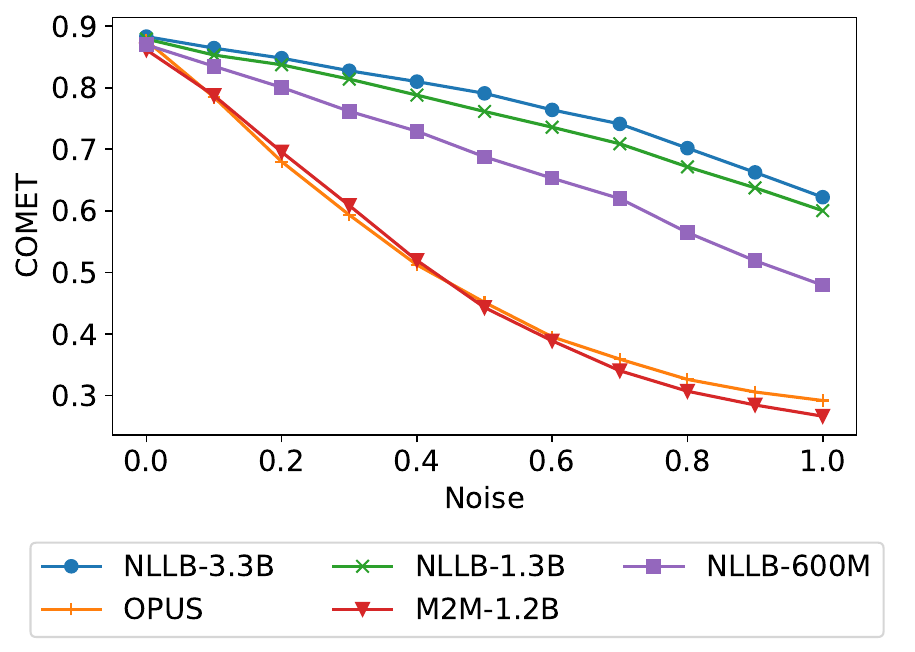}
    \caption{COMET on \langp{en}{fr} swaps.}
    \label{fig:size-comet}
\end{figure}

\paragraph{Size and multilinguality.}
From these experiments, one might conclude that robustness depends largely on model size (\opus{} is $14$ times smaller than any other system) or multilinguality (all except \opus{} are multilingual).
However, Figure~\ref{fig:size-comet} shows that these are not the only factors.
We reran swap experiments with NLLB-600M, NLLB-1.3B,
and M2M-1.2B \citep{fan2021beyond};
despite the similar sizes of \textsc{NLLB-1.3B} and \textsc{M2M-1.2B}, and the fact that they are both massively multilingual, they handle noise differently: \textsc{NLLB-1.3B} is similar to \textsc{NLLB-3.3B}, while \textsc{M2M-1.2B} suffers as much as \opus.

\paragraph{Impact of architecture.}
Having shown that model size does not have a strong effect on robustness (at least not for \nllb{}), we next investigate the impact of architecture on performance.
Is the gap between \opus{} and other models primarily due to differences in training data, or is there some aspect of the LLMs' decoder-only structure that encourages robustness?
To investigate, we trained a 1.3B parameter\footnote{We would have preferred to train a 13B model (similar to \ti), but this was impossible due to resource constraints.} decoder-only model on the same Tatoeba Challenge data as was used by the \langp{en}{pt} \opus{} model.
Training details are given in Appendix~\ref{appendix:opusllm}.
The performance and robustness of this model, which we dub OPUSLLM, are shown in Table \ref{table:opusllm}.
Although its performance on clean data lags $3$ COMET points behind \opus{}, the COMET-slope is similar for all four noise types, suggesting that the robustness of recent models is due to their training data, not their size or architecture.

\paragraph{Tokenizer robustness.}
Introducing perturbations affects not only translation quality but also runtime.
Perturbations create character sequences that are less similar to the data that tokenizers are trained on, which leads to more pieces being used to encode the sentence.
This is true even for drop noise, which increases the length of the tokenized sequence even as it \textit{shortens} the detokenized sequence.
In Table~\ref{table:fertility}, we compare tokenizers by their fertility --- the average number of subword pieces per whitespace word --- on clean and key data.
While \opus{} tokenizers generally have very low fertility on clean data, it increases more than the other tokenizers, suggesting the tokenizer itself is less robust to character perturbations.
It is also notable that \ti{} and \cgpt{} have high fertility even on clean Korean text.
While this is a symptom of tokenizer unfairness in large models \citep{petrov2023language}, it can also be a sign of tokenizer robustness: the higher the fertility, the closer the model is to byte-level tokenization.
This results in noisy token sequences that are much closer to the clean sequences for \ti{} and \cgpt{}, as can be seen in terms of F1 in Table~\ref{table:korean-tokens}.
The same trend does not hold for the other languages.

\begin{table}[t]
    \fontsize{9pt}{11pt}\selectfont
    \centering
    \begin{tabular}{lrrrrrr}
        \toprule
         & \multicolumn{2}{c}{English} & \multicolumn{2}{c}{Portuguese} & \multicolumn{2}{c}{Korean} \\
         & clean & key & clean & key & clean & key \\
         \midrule
         \opus & 1.25 & 2.71 & 1.29 & 2.62 & 1.75 & 2.93\\
         \nllb & 1.40 & 2.20 & 1.53 & 2.21 & 2.03 & 3.02\\
         \ti   & 1.42 & 2.47 & 1.88 & 2.55 & 6.37 & 7.44\\
         \cgpt & 1.24 & 2.23 & 1.71 & 2.34 & 4.17 & 5.12\\
         \bottomrule
    \end{tabular}
    \caption{Tokenizer fertility with clean and key-perturbed data. For English, we used the \langp{en}{fr} \opus{} model.}
    \label{table:fertility}
\end{table}

\begin{table}[t]
    \small
    \centering
    \begin{tabular}{lrrrr}
    \toprule
    & swap & drop & dupe & key \\
    \midrule
    \opus  & 21.6 & 27.3 & 36.9 & 33.8\\
    \nllb  & 27.6 & 35.1 & 45.7 & 42.3\\
    \ti    & 50.0 & 62.4 & 74.8 & 71.9\\
    \cgpt  & 39.5 & 52.3 & 65.5 & 62.8\\
    \bottomrule
    \end{tabular}

    \caption{F1 between clean Korean token sequences and their 100\% noisy counterparts.}
    \label{table:korean-tokens}
\end{table}
\section{Robustness to Social Media Text}

The previous experiments show that large translation models and LLMs are more robust to synthetic character perturbations than conventional MT models.
But is this result applicable to ``authentically noisy'' domains such as social media text?
The nature of ``noise'' here is different than in the synthetic task: social media text does not necessarily contain many errors \citep{rello2012social}, but the domain is very different from \flores.
This makes it difficult to isolate the effect of noise from the general domain adaptation problem.
Ideally, we would have a translation corpus in which each example is a triple consisting of an original noisy source, a manually annotated cleaned source, and a gold standard translation.
This would allow translations of clean and noisy versions of the same source to be compared on some reference-based metric, isolating the effect of the errors.
As we are aware of only one such corpus \citep{bawden-sagot-2023-rocs},
we instead perform two complementary investigations.
First, we evaluate on MTNT \citep{michel-neubig-2018-mtnt}, a noisy social media MT corpus.
Although this is a useful test of our models, the noise is not labeled and there is no clean version of the same data to compare to.
This motivates our second experiment, in which we translate data from MultiLexNorm \citep{van-der-goot-etal-2021-multilexnorm}, a lexical normalization benchmark.
Together, these experiments allow us to see both which models succeed and how badly they fail.

\subsection{MTNT Experiments}
\label{subsec:mtnt-exp}
MTNT pairs Reddit posts with high-quality professional translations.
Although the references are somewhat clean, the sources are only lightly filtered, making them potentially noisy.\footnote{$2.18\%$ percent of MTNT \langp{en}{fr} source tokens are misspelled \citep{michel-neubig-2018-mtnt}.
This is a higher rate than in formal corpora, but lower than in our synthetic experiments.}
Unfortunately no cleaned sources exist, making the effect of noise difficult to isolate.
Despite this difficulty, it is often used as a robustness benchmark \citep[\it inter alia]{karpukhin-etal-2019-training,park-etal-2020-adversarial,salesky-etal-2021-robust,vaibhav-etal-2019-improving}.

\paragraph{Finetuning.}
We finetuned \opus{} on MTNT \langpb{en}{fr} as described in Appendix~\ref{appendix:training}.
We dub this model r/\opus.

\begin{table}[t]
    \small
    \centering
\begin{tabular}{lrr}
\toprule
Method & \langp{en}{fr} & \langp{fr}{en}\\
\midrule
\opus      & 77.21 & 79.64\\
r/\opus  & 79.22 & 81.94\\
\nllb      & 79.33 & 80.59\\
\ti       & \textbf{81.91} & 83.66\\
\cgpt & 81.33 & \textbf{84.72}\\
\bottomrule
\end{tabular}
    \caption{COMET on the MTNT test set.}
    \label{table:mtnt-test}
\end{table}

\paragraph{Results.}
Results are shown in Table~\ref{table:mtnt-test}.
Despite in-domain finetuning benefiting \opus{} by more than 2 COMET points for both \langp{en}{fr} and \langp{fr}{en}, this does not close the gap to \ti{} and \cgpt{}. 
This suggests that \ti{} and \cgpt{} benefit from their massive training corpora, which likely contain large quantities of social media text.
In contrast, the only social media text the finetuned \opus{} models have seen are MTNT's tiny training sets (36k parallel examples for \langp{en}{fr}, 19k for \langp{fr}{en}), plus whatever is in the Tatoeba Challenge corpora.

\begin{table*}[t]
    \small
    \centering
\begin{tabular}{l|rrr|rrr|rrr|rrr}
\toprule
& \multicolumn{3}{c}{\langp{en}{de}} & \multicolumn{3}{c}{\langp{de}{en}} & \multicolumn{3}{c}{\langp{en}{es}} & \multicolumn{3}{c}{\langp{es}{en}}\\
Model & FB & FC & \dqe & FB & FC & \dqe & FB & FC & \dqe & FB & FC & \dqe \\
\midrule
\opus    & 81.9 & 81.50 & 6.21 & 74.8 & 88.36 & 3.95 & 80.3 & 82.54 & 4.11 & 86.2 & 88.52 & 2.54\\
\nllb    & 87.3 & 87.28 & 2.81 & 74.8 & 87.75 & 2.88 & 88.3 & 88.62 & 2.24 & 84.2 & 88.70 & 2.30\\
\ti   & \textbf{88.8} & \textbf{89.07} & 1.86 & 75.2 & 90.57 & 2.41 & 89.0 & \textbf{90.26} & 1.19 & 85.3 & 89.32 & 2.63\\
\cgpt    & 87.1 & 88.45 & \textbf{1.15} & \textbf{83.1} & \textbf{91.86} & \textbf{1.32} & \textbf{89.1} & 90.14 & \textbf{0.72} & \textbf{87.9} & \textbf{91.23} & \textbf{0.91}\\
\bottomrule
\end{tabular}
    \caption{MultiLexNorm results. FB is faux-BLEU and FC is faux-COMET.}
    \label{table:multilexnorm-cometkiwi}
\end{table*}

\subsection{MultiLexNorm Experiments}
\label{subsec:multilexnorm-exp}

While MTNT is an established benchmark and useful sanity check, it is not controllable like our synthetic experiments; we cannot isolate the effect of noise because there is no non-noisy version of the corpus.
Therefore we pivot to evaluate models on translating MultiLexNorm \citep{van-der-goot-etal-2021-multilexnorm}, a lexical normalization dataset that pairs social media text primarily from Twitter with manually cleaned versions of the same.
Switching from MTNT to MultiLexNorm comes with a trade-off: in order to gain clean sources, we lose references.

\paragraph{Data.}
We use the English, German, and Spanish data from MultiLexNorm as our translation sources.
In experiments with English sources, we translate to German and Spanish; otherwise, we translate to English.
Statistics are presented in Appendix~\ref{appendix:multilexnorm}.

\paragraph{Evaluation.}
As MultiLexNorm lacks reference translations, we use three reference-free metrics.
Faux-BLEU \citep{anastasopoulos-2019-analysis} computes $\text{spBLEU}(y_n, y_c)$ \citep{papineni-etal-2002-bleu,nllbteam2022language}, 
where $y_n$ and $y_c$ are the hypotheses computed from the noisy source and the clean source, respectively.
$y_c$ is treated as a pseudoreference.
By analogy we also compute faux-COMET.
These faux-metrics measure the \textbf{similarity} between $y_c$ and $y_n$, with faux-BLEU capturing lexical features while faux-COMET is ``deeper.''
In addition, we use a new metric that we dub \dqe{}.
Given $y_n$, $y_c$, noisy and clean source sequences $x_n$ and $x_c$, and a reference-free quality estimation metric $\mathqe$, $\mathdqe = \mathqe(x_c, y_c) - \mathqe(x_c, y_n)$.
We compute $\mathqe$ with COMETKiwi\footnote{Specifically, we use \texttt{Unbabel/wmt22-cometkiwi-da}.} \citep{rei-etal-2022-cometkiwi}.
A \dqe{} close to zero means that a model produces similar-quality outputs for both inputs, indicating robustness, whereas a large positive value indicates that translation quality suffers on noisy data.

\paragraph{Results.}
Table~\ref{table:multilexnorm-cometkiwi} shows the performance of all models with both noisy and gold-standard cleaned versions of the corpora.
In terms of \dqe{}, \cgpt{} performs best for all language pairs.
It also records the best faux-BLEU for all pairs except \langp{en}{de}, on which \ti{} and \nllb{} both outperform it.
The faux-COMET results show a split, where \cgpt{} has the highest scores for both to-English pairs but \ti{} passes it for \langp{en}{de} and \langp{en}{es}.
\section{Strategies for Mitigating Noise}
\label{section:correction}

So far we have shown that \opus{} is less robust to synthetic noise than larger models and performs worse on social media text.
Next we evaluate two techniques for mitigating noise: finetuning MT models on synthetically noised data and incorporating a source correction pipeline.
These approaches have contrasting trade-offs.
Finetuning the MT model allows the robust translation task to be learned end-to-end.
However, this is expensive if a model is very large (\ti, \nllb) and impossible if its weights are closed (\cgpt).
On the other hand, pipelines are modular, allowing the same correction system to be reused with any model.
The drawback of this modularity is that pipelines may introduce errors as well as fixing them.

\subsection{Synthetic Experiments}\label{subsec:synth-mitigation}

As a comparison of finetuning and source correction, we focus on \langp{en}{pt} with the same synthetically noised corpora as in Section~\ref{section:synth-robustness}.
We kept the training data as similar as possible between finetuning and correction experiments: in both cases, we subsampled 5 million examples from the Tatoeba Challenge \langp{en}{pt} training set.
We noised 15\% of source tokens with each of the four noise types used in Section \ref{section:synth-robustness}, so a total of 60\% of tokens were corrupted.
For validation, we concatenated the original \flores{} \langp{en}{pt} development set to a version of the same set in which 20\% of tokens have been corrupted with each noise type.
By including clean validation data, this favors models that do not forget how to translate clean data.
For finetuning \opus{}, these noised source examples are paired with clean Portuguese targets.
For finetuning correction models, they are paired with the original clean English sources.
We followed the training procedure in Appendix~\ref{appendix:training} for both MT and correction models.

\begin{table}[t]
    \small
    \centering
    \begin{tabular}{lrrrrrr}
    \toprule
        Model    & clean & swap & drop & dupe & key\\
        \midrule
        \opus & 88.94 & -72.97 & -69.66 & -57.94 & -75.81\\
        +finetuning & 88.52 & -2.14 & -7.59 & -0.87 & -5.01\\
        +correction & 88.36 & -2.02 & -11.81 & -0.08 & -7.20\\
        \midrule
        \nllb       & 89.58 & -22.41 & -21.71 & -4.13 & -25.57\\
        +correction & 89.11 &  -1.57 & -9.09  & -0.07 & -5.47\\
        \midrule
        \ti      & 90.02 & -13.44 & -14.68 & -1.89 & -22.04\\
        +correction & 89.57 &  -0.86 &  -5.59 & -0.16 & -2.99\\
        \midrule
        ChatGPT & 89.83 & -3.76 & -6.63 & -0.98 & -7.78\\
        \bottomrule
    \end{tabular}
    \caption{Clean COMET and COMET-slope for \langp{en}{pt} with finetuning and source correction.}
    \label{table:synthetic-mitigation}
\end{table}

\begin{table}[t]
    \small
    \centering
    \textbf{MT Finetuning}\\
    \begin{tabular}{rrrrrrr}
    \toprule
        Model    & clean & swap & drop & dupe & key\\
        \midrule
        100k & 88.28 & -12.08 & -18.96 & -4.63 & -18.65\\
        1m  & 88.13   & -5.21 & -11.72 & -2.04 & -9.49\\
        5m & 88.52   & -2.14  &  -7.59 & -0.87 & -5.01\\
        \bottomrule
    \end{tabular}

    \bigskip
    
    \textbf{Source Correction}\\
    \begin{tabular}{rrrrrrr}
    \toprule
        Model    & clean & swap & drop & dupe & key\\
        \midrule
        100k & 88.59 & -4.12 & -15.31 & -0.11 & -10.84\\
        1m   & 88.30 & -2.25 & -12.83 & -0.15 & -7.83\\
        5m   & 88.36 & -2.02 & -11.81 & -0.08 & -7.20\\
        \bottomrule
    \end{tabular}
    \caption{Performance of noise mitigation with \opus{} \langp{en}{pt} at varying quantities of noisy training data.}
    \label{table:training-size}
\end{table}

\begin{figure}
    \centering
    \includegraphics[width=1.0\linewidth]{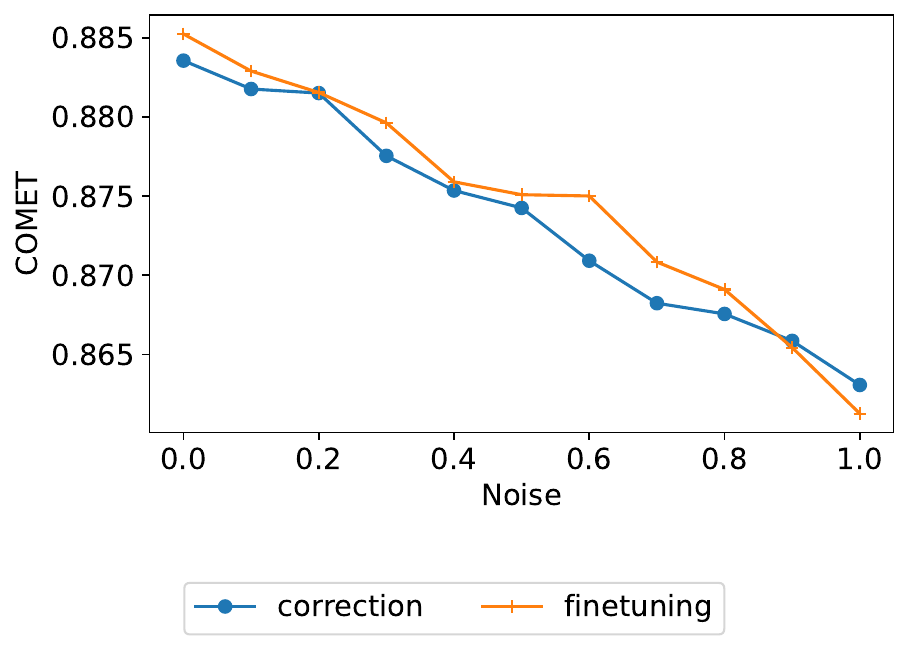}
    \caption{\opus{} \langp{en}{pt} swaps with finetuning and SC.}
    \label{fig:corr-finetune-swap}
\end{figure}

\paragraph{Source correction (SC).}
As our base model for SC, we adopted ByT5-Small \citep{xue-etal-2022-byt5}.
At inference time we report the results of a pipeline that pairs the corrector with a translation model.
In intrinsic terms, this corrector manages a chrF \citep{popovic-2015-chrf} of at least $89.6$ at correcting each noise type, and over $97.5$ for swaps and dupes.
Additional results are shown in Appendix~\ref{appendix:correction}.

\begin{figure}[t]
    \centering
    \includegraphics[width=1.0\linewidth]{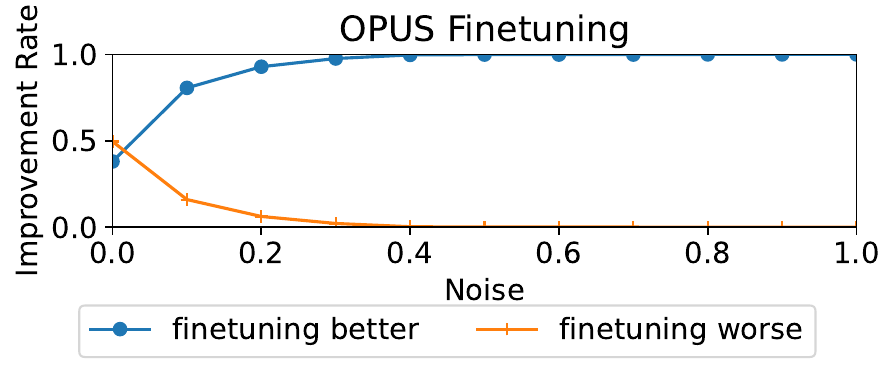}

    \includegraphics[width=1.0\linewidth]{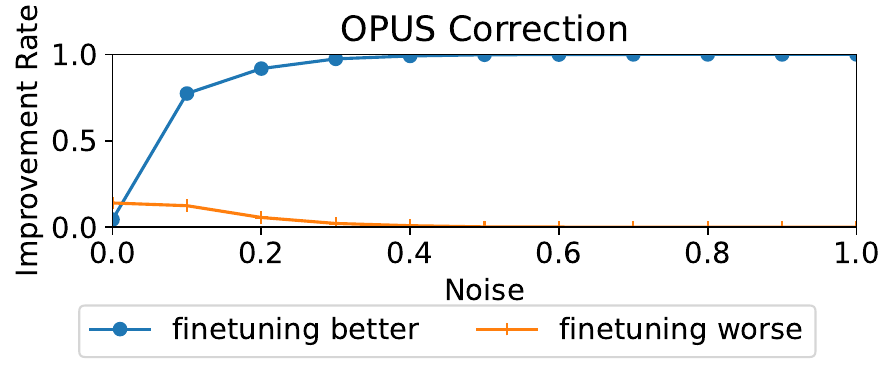}

    \includegraphics[width=1.0\linewidth]{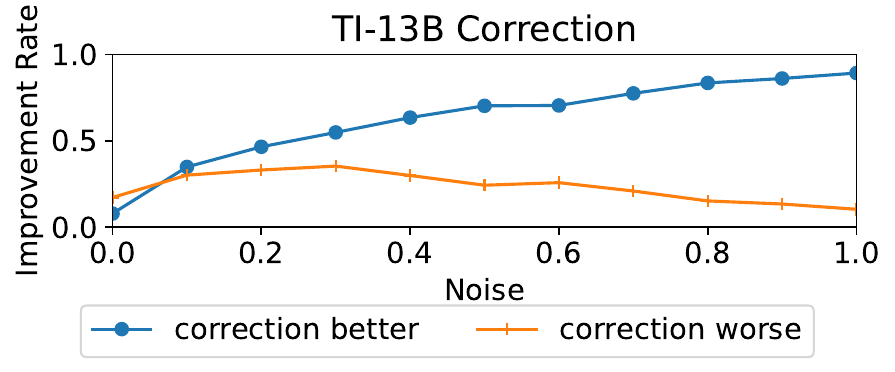}

    \caption{Percentage of \langp{en}{pt} swap examples for which finetuning \opus{} (top), correcting \opus{} (middle), or correcting \ti{} (bottom) outperforms the baseline.}
    \label{fig:corr-finetuning-improvement}
\end{figure}

\paragraph{Results.}
Our main results are shown in Table~\ref{table:synthetic-mitigation}.
For \opus{}, both MT finetuning and correction greatly improve robustness to synthetic errors, with both techniques reporting flatter COMET-slopes than \cgpt{} for all noise types except drop.
At a more granular level, in Figure~\ref{fig:corr-finetune-swap} we show \opus's COMET scores at all levels of swap noise for MT finetuning and SC.
It is clear that MT finetuning outperforms SC across almost all noise levels.
Despite \nllb{} and \ti{} being more robust than \opus{} in general, they too benefit from SC, suggesting that its effect is complementary with the models' inherent robustness.
Both MT finetuning and SC slightly degrade model performance on clean translation.
However, these degradations are small, and become smaller as the base model gets larger: corrected \opus{} declines $0.58$ COMET, versus only $0.45$ for \ti{}.
These results also show the surprising difficulty of drop errors: although these are not more problematic for baseline models than other error types (see Tables~\ref{table:flores-slope-xx-en} and \ref{table:flores-slope-en-xx}), neither correction nor MT finetuning handles them as effectively as the other error types.
This suggests that the missing information from a single deleted character often cannot be recovered from surrounding context.

\paragraph{Effect of training size.}
Although we used 5 million training examples in our main experiments, Table~\ref{table:training-size} shows the effect that using fewer has on \opus's robustness.
The correction approach performs much better when data is restricted to 100k examples, but this advantage shrinks as the training size is increased and eventually MT finetuning outperforms it.
This supports the intuition that noisy translation is a harder task than correction, so the \opus{} model requires more examples to learn it.

\paragraph{How often does mitigation work?}
In the main results, we showed that MT finetuning and SC both greatly improve robustness.
However, as the scores are presented at corpus level, they do not give insight into what percentage of examples are actually helped, versus how many are harmed.
To fill in this missing piece, we compared sentence-level COMET between our baseline models and the two mitigation approaches.
Results are shown in Figure~\ref{fig:corr-finetuning-improvement}.
At high noise levels, both techniques nearly always help \opus{}.
However, they behave differently on clean sources.
While SC almost never improves clean scores and harms less than 10\% of examples, finetuning is more of a high-risk strategy, as it makes scores worse for about half of examples while also improving COMET for 38\% of them.
The trend is subtly different for SC with \ti.
While correction helps more examples than it hurts whenever the noise level is at least 10\%, the percentage of examples that are harmed by correction actually \emph{increases} up to 30\% noise.
This suggests that correction may introduce errors into some noisy sequences that \ti{} could have handled itself.

\subsection{Mitigating Errors in MTNT}
Having shown that MT finetuning and SC are both effective techniques for improving robustness to synthetic errors, we return to MTNT.
Our goal is to determine whether either MT finetuning or SC can provide some benefit to performance on \langp{en}{fr}.

\begin{table}[t]
    \small
    \centering
\begin{tabular}{lrr}
\toprule
Finetuning & Uncorrected & Corrected\\
\midrule
None     & 77.21 & 76.31\\
MTNT     & \textbf{79.22} & 77.97\\
Synthetic & 75.95 & 75.31\\
\bottomrule
\end{tabular}
    \caption{\opus{} performance on the MTNT \langp{en}{fr} test set with various finetuning techniques.}
    \label{table:mtnt-opus-mitigation}
\end{table}

\paragraph{Finetuning vs. correction for \opus.}
We evaluate MT finetuning and SC on MTNT.
For MT finetuning, we compare two techniques: finetuning only on MTNT, as in Section~\ref{subsec:mtnt-exp};
and finetuning on 5 million synthetically noised Tatoeba Challenge \langp{en}{fr} examples, matching the procedure used for \langp{en}{pt} in Section~\ref{subsec:synth-mitigation}.
For SC, we use the same English model as in the synthetic experiments.
Results are shown in Table~\ref{table:mtnt-opus-mitigation}.
Finetuning on MTNT outperforms the baseline, but both SC and MT finetuning are harmful, confirming that MTNT is not very noisy in terms of spelling errors \citep{karpukhin-etal-2019-training,michel-neubig-2018-mtnt}.

\paragraph{Effect of source correction.}
In Table~\ref{table:mtnt-correction}, we show results with and without SC for various models.
Although correction does not improve results on the whole, the oracle consistently outperforms the baseline by about $0.5$ COMET, illustrating that many sequences do benefit from correction.
Indeed, 21.8\% of MTNT examples improve with correction and another 46.1\% are not harmed.
Future work could investigate the use of a routing mechanism that decides for each example whether to apply SC.

\begin{table}[t]
    \small
    \centering
    \begin{tabular}{lrrr}
    \toprule
        Method    & r/\opus & \nllb & \ti\\
        \midrule
        Base      & 79.22 & 79.33 &  81.91\\
        \midrule
        Correction & 77.97 & 78.32 & 80.94\\
        Oracle     & 79.71 & 79.82 & 82.43\\
        \bottomrule
    \end{tabular}
    \caption{MTNT results with and without source correction. The oracle selects the translation with the higher COMET between the baseline and the pipeline.}
    \label{table:mtnt-correction}
\end{table}

\section{Conclusion}
We presented several experiments testing the robustness of MT systems to synthetic and natural noise.
On synthetic noise, we showed that large multilingual MT models and LLMs are far more robust than older models.
The experiments on social media translation showed that larger models also worked better on natural noise.
We further supported this conclusion through reference-free translation experiments with a novel evaluation metric based on quality estimation.
Finally, we showed that noisy finetuning and source correction allow smaller models to exceed GPT-3.5's robustness synthetic noise, while also being useful in many cases for handling natural errors.

\section*{Limitations}
We acknowledge the limitations of our work.
All of the languages included in this study have large speaker populations and many resources available, and are the official languages of countries.
Some conclusions may not generalize to low resource languages.
Additionally, this paper studies only one source of natural noise, namely social media text.
Other varieties of text perceived as noisy, such as transcribed speech or text written by nonfluent language users, may have different properties.
Finally, the evaluation techniques used in this work are all automatic or neural, and may differ from gold-standard human evaluation.

\section*{Acknowledgments}
This work was supported by EU's Horizon Europe Research and Innovation Actions (UTTER, contract 101070631), by the project DECOLLAGE (ERC-2022-CoG 101088763), by the Portuguese Recovery and Resilience Plan through project C645008882-00000055 (Center for Responsible AI), and by FCT/MECI through national funds and when applicable co-funded EU funds under UID/50008: Instituto de Telecomunicações.
We thank Sweta Agrawal and the anonymous reviewers for their helpful comments.

\bibliography{anthology_p1,anthology_p2,custom}

\appendix

\section{Key Noise}
\label{appendix:key-noise}

As different languages customarily use different keyboard layouts, we made slight alterations to our key noising procedure for each source language.
We use the QWERTZ layout for German, AZERTY for French, QWERTY for English and Portuguese, and South Korean Dubeolsik for Korean.
For Korean, we used \texttt{hangul-jamo}\footnote{\url{https://github.com/jonghwanhyeon/hangul-jamo}} to decompose hangul characters into jamo, which represent individual keystrokes, before applying perturbations.

\section{OPUS Models}
\label{appendix:opus-model}

\begin{table}[ht]
    \centering
    \begin{tabular}{lrr}
    \toprule
        LP & Vocab & Params\\
        \midrule
        \langp{en}{de} & 58.1k & 74.4M\\
        \langp{de}{en} & 58.1k & 74.4M\\
        \langp{en}{es} & 55.0k & 234.8M\\
        \langp{es}{en} & 65.0k & 77.9M\\
        \langp{en}{fr} & 53.0k & 232.7M\\
        \langp{fr}{en} & 53.0k & 232.7M\\
        \langp{en}{ko} & 32.0k* & 209.2M\\
        \langp{ko}{en} & 32.0k* & 209.2M\\
        \langp{en}{pt} & 54.7k & 234.5M\\
        \langp{pt}{en} & 60.0k & 75.4M\\
        \bottomrule
    \end{tabular}
    \caption{\opus{} models. Each path is preceded by \texttt{Helsinki-NLP/}. *\langpb{en}{ko} use separate 32k source and target vocabularies. All others use shared vocabularies.}
    \label{tab:opus-models}
\end{table}

\noindent The parameters and vocabulary sizes of the \opus{} models are shown in Table~\ref{tab:opus-models}.
All checkpoints come from the Tatoeba Challenge \citep{tiedemann-2020-tatoeba}.\footnote{\url{https://github.com/Helsinki-NLP/Tatoeba-Challenge/tree/master/models}}

\section{Inference}
\label{appendix:inference}

Owing to the differing frameworks between \opus, OPUSLLM, \nllb, \ti{}, and ByT5 models, we use different beam search implementations depending on the model.
For \opus{} models, we decode with Marian \citep{junczys-dowmunt-etal-2018-marian}, while \nllb{} and ByT5 use Hugging Face \texttt{transformers}.\footnote{\url{https://github.com/huggingface/transformers}}
\ti{} models and OPUSLLM are decoded with \texttt{vllm} \citep{kwon2023efficient}.\footnote{\url{https://github.com/vllm-project/vllm}}
Regardless of framework, we used a beam size of $5$ across all experiments.

\subsection{Prompt for TowerInstruct}

\texttt{``Translate the following text from \textbf{[source language]} to \textbf{[target language]}.$\backslash$nSource:\textbf{[source text]}$\backslash$n\textbf{[target language]}:''}

\subsection{Prompt for GPT-3.5}

\texttt{``Translate this sentence from \textbf{[source language]} to \textbf{[target language]}.$\backslash$nSource:\textbf{[source text]}$\backslash$nTarget:''}

\section{OPUSLLM Training Details}
\label{appendix:opusllm}
The OPUSLLM model was trained with used Megatron-DeepSpeed\footnote{\url{https://github.com/microsoft/Megatron-DeepSpeed}} for a maximum of 300k steps with an effective batch size of 65k tokens and a base learning rate of $3 \times 10^{-4}$, with a constant learning rate schedule and 5000 warmup steps.
The model used the same 32k tokenizer as \ti.
To account for the model's decoder-only structure, examples were formatted with two special tokens to indicate the beginning of the source and target sequences.
At inference time, we selected the checkpoint with the best COMET on the FLORES \langp{en}{pt} dev set.

\section{Training Hyperparameters}
\label{appendix:training}
In this work, we finetune two types of base models: \opus{} models and ByT5-Small.
Despite the various technical differences between these two models, we used almost identical training procedures for them, with differences noted below.

\paragraph{Basic training procedure.}
We finetuned using early stopping with patience 3. 
We validated every 500 steps.
We used a grid over the learning rates $\{10^{-4}, 10^{-5}, 10^{-6}\}$ and selected the best checkpoint by validation loss.

\paragraph{Training library.}
For \opus{} models, we finetuned with the \texttt{marian} command line tool from Marian \citep{junczys-dowmunt-etal-2018-marian}.
For ByT5-Small, we used a script that leverages the \texttt{Trainer} class from Hugging Face \texttt{transformers}.

\section{MultiLexNorm Statistics}
\label{appendix:multilexnorm}

\begin{table}[ht]
    \small
    \centering
    \begin{tabular}{lrrl}
    \toprule
        Lang. & Sent. & \%Noisy & Reference \\
        \midrule
        English  & 1967 & 6.9 &\citep{baldwin-etal-2015-shared} \\
        German   & 583  & 8.9 &\citep{sidarenka2013rule} \\
        Spanish & 531   & 7.7 &\citep{alegria2013introduccion}\\
        \bottomrule
    \end{tabular}
    \caption{Statistics of selected MultiLexNorm corpora.}
    \label{tab:multilexnorm-data}
\end{table}

\section{Intrinsic Correction Performance}
\label{appendix:correction}

\begin{table}[ht]
    \small
    \centering
    \begin{tabular}{lrrrr}
    \toprule
    & swap & drop & dupe & key \\
    \midrule
    None       & 40.0 & 47.0 & 68.2 & 45.7\\
    JamSpell   & 75.5 & 63.5 & 91.7 & 78.0\\
    ByT5-Small & 97.5 & 89.6 & 99.6 & 94.4\\
    \bottomrule
    \end{tabular}

    \caption{chrF \citep{popovic-2015-chrf} of source correctors at 100\% noise in \flores{}.}
    \label{table:corrector-intrinsic}
\end{table}

Intrinsic correction results in terms of chrF \citep{popovic-2015-chrf} are shown in Table~\ref{table:corrector-intrinsic}.
In addition to our ByT5-Small corrector, we include the results of the pretrained English model from JamSpell\footnote{\url{https://github.com/bakwc/JamSpell}} \citep{jamspell}, a trigram-based spell-checker.

\section{Synthetic Results with Other Metrics}
\label{appendix:synth-metrics}
The models' synthetic performance on clean data is shown in terms of BLEU\footnote{\texttt{nrefs:1|case:mixed|eff:no|tok:flores200|\\smooth:exp|version:2.5.1}} in Table~\ref{table:clean-flores-bleu} and in chrF\footnote{\texttt{nrefs:1|case:mixed|eff:yes|nc:6|nw:0|space:no|\\version:2.5.1}} in Table~\ref{table:clean-flores-chrf}.
Synthetic results showing BLEU-slope are shown in Tables~\ref{table:flores-bleu-slope-xx-en} and \ref{table:flores-bleu-slope-en-xx}.
Synthetic results showing chrF-slope are shown in Tables~\ref{table:flores-chrf-slope-xx-en} and \ref{table:flores-chrf-slope-en-xx}.
BLEU-slope and chrF-slope are computed analogously to COMET-slope.

\begin{table}[t]
    \centering
    \small
\textbf{\langp{xx}{en}}\\
\begin{tabular}{lrrrrr}
\toprule
Model & \langp{de}{en} & \langp{fr}{en} & \langp{ko}{en} & \langp{pt}{en} & avg.\\
\midrule
\opus    & 44.7 & 49.7 & 30.7 & 49.3 & 43.6\\
\nllb    & 49.3 & 50.3 & 34.8 & 55.1 & 47.4\\
\ti      & \textbf{50.6} & \textbf{52.1} & \textbf{37.0} & \textbf{56.7} & \textbf{49.1}\\
\cgpt    & 48.8 & 49.1 & 32.5 & 53.6 & 46.0\\
\bottomrule
\end{tabular}

\bigskip

\textbf{\langp{en}{xx}}\\
\begin{tabular}{lrrrrr}
\toprule
Model & \langp{en}{de} &  \langp{en}{fr} &  \langp{en}{ko} & \langp{en}{pt} & avg. \\
\midrule
\opus    & 42.0 & 55.6 & 23.8 & 54.6 & 44.0\\
\nllb    & 46.8 & 56.2 & 24.8 & 54.5 & 45.6\\
\ti      & 47.0 & \textbf{57.1} & \textbf{29.5} & 52.8 & 46.6\\
\cgpt    & \textbf{47.8} & 56.6 & 25.6 & \textbf{56.6} & \textbf{46.7}\\
\bottomrule
\end{tabular}
    \caption{BLEU on \flores{} without added noise.}
    \label{table:clean-flores-bleu}
\end{table}

\begin{table}[t]
    \centering
    \small
\textbf{\langp{xx}{en}}\\
\begin{tabular}{lrrrrr}
\toprule
Model & \langp{de}{en} & \langp{fr}{en} & \langp{ko}{en} & \langp{pt}{en} & avg.\\
\midrule
\opus    & 66.4 & 69.1 & 55.8 & 69.4 & 65.2\\
\nllb    & 69.2 & 69.9 & 58.6 & 72.9 & 67.7\\
\ti      & \textbf{69.9} & \textbf{70.6} & \textbf{60.3} & \textbf{73.9} & \textbf{68.7}\\
\cgpt    & 69.8 & 69.7 & 58.2 & 72.7 & 67.6\\
\bottomrule
\end{tabular}

\bigskip

\textbf{\langp{en}{xx}}\\
\begin{tabular}{lrrrrr}
\toprule
Model & \langp{en}{de} &  \langp{en}{fr} &  \langp{en}{ko} & \langp{en}{pt} & avg. \\
\midrule
\opus    & 62.9 & 71.7 & 36.2 & 71.3 & 60.5\\
\nllb    & 65.7 & 71.8 & 36.7 & 71.0 & 61.3\\
\ti      & 66.2 & 72.3 & \textbf{38.7} & 70.3 & \textbf{61.9}\\
\cgpt    & \textbf{67.1} & \textbf{72.6} & 35.1 & \textbf{72.8} & \textbf{61.9}\\
\bottomrule
\end{tabular}
    \caption{chrF on \flores{} without added noise.}
    \label{table:clean-flores-chrf}
\end{table}

\begin{table}[t]
    \small
    \centering

\textbf{swap}\\
\begin{tabular}{lrrrrr}
\toprule
Model & \langp{de}{en} & \langp{fr}{en} & \langp{ko}{en} & \langp{pt}{en} & avg.\\
\midrule
\opus  & -51.4 & -58.2 & -28.2 & -56.2 & -48.5 \\
\nllb  & -33.4 & -35.8 & -30.6 & -37.5 & -34.3 \\
\ti    & -37.9 & -38.8 & -30.4 & -39.6 & -36.7 \\
\cgpt  & \textbf{-10.9} & \textbf{-14.4} & \textbf{-27.4} & \textbf{-17.2} & \textbf{-17.5} \\
\bottomrule
\end{tabular}

\textbf{drop}\\
\begin{tabular}{lrrrrr}
\toprule
Model & \langp{de}{en} & \langp{fr}{en} & \langp{ko}{en} & \langp{pt}{en} & avg.\\
\midrule
\opus  & -45.6 & -49.6 & -26.4 & -49.1 & -42.7\\
\nllb  & -29.8 & -29.9 & -26.1 & -32.7 & -29.6\\
\ti    & -29.5 & -29.7 & -25.6 & -30.0 & -28.7\\
\cgpt  & \textbf{-14.4} & \textbf{-13.8} & \textbf{-24.9} & \textbf{-16.0} & \textbf{-17.3}\\
\bottomrule
\end{tabular}

\textbf{dupe}\\
\begin{tabular}{lrrrrr}
\toprule
Model & \langp{de}{en} & \langp{fr}{en} & \langp{ko}{en} & \langp{pt}{en} & avg.\\
\midrule
\opus  & -34.3 & -38.0 &  -8.8 & -37.6 & -29.7\\
\nllb  & -13.9 & -15.7 & -11.5 & -14.8 & -14.0\\
\ti    &  -9.9 & -13.0 &  \textbf{-5.5} & -10.5 & -9.7\\
\cgpt  &  \textbf{-5.5} &  \textbf{-5.3} & -12.0 &  \textbf{-5.8} & \textbf{-7.2}\\
\bottomrule
\end{tabular}

\textbf{key}\\
\begin{tabular}{lrrrrr}
\toprule
Model & \langp{de}{en} & \langp{fr}{en} & \langp{ko}{en} & \langp{pt}{en} & avg.\\
\midrule
\opus  & -50.3 & -57.3 & -27.8 & -57.2 & -48.2\\
\nllb  & -35.2 & -36.4 & -33.3 & -40.3 & -36.3\\
\ti    & -38.5 & -42.7 & -27.2 & -46.3 & -38.7\\
\cgpt  & \textbf{-18.3} & \textbf{-18.2} & \textbf{-22.5} & \textbf{-21.9} & \textbf{-20.2}\\
\bottomrule
\end{tabular}

    \caption{BLEU-slope on \flores{} for \langp{xx}{en}.}
    \label{table:flores-bleu-slope-xx-en}
\end{table}

\begin{table}[t]
    \small
    \centering

\textbf{swap}\\
\begin{tabular}{lrrrrr}
\toprule
Model & \langp{en}{de} & \langp{en}{fr} & \langp{en}{ko} & \langp{en}{pt} & avg.\\
\midrule
\opus  & -49.1 & -62.6 & -28.9 & -64.5 & -51.3\\
\nllb  & -31.2 & -33.8 & -18.1 & -37.1 & -30.0\\
\ti    & -19.8 & -20.7 & -12.6 & -20.0 & -18.3\\
\cgpt  &  \textbf{-7.8} &  \textbf{-9.3} &  \textbf{-7.1} &  \textbf{-5.1} &  \textbf{-7.3}\\
\bottomrule
\end{tabular}

\textbf{drop}\\
\begin{tabular}{lrrrrr}
\toprule
Model & \langp{en}{de} & \langp{en}{fr} & \langp{en}{ko} & \langp{en}{pt} & avg.\\
\midrule
\opus  & -47.8 & -59.4 & -27.9 & -61.4 & -49.1\\
\nllb  & -30.7 & -33.7 & -16.6 & -36.2 & -29.3\\
\ti    & -21.9 & -23.5 & -13.7 & -22.6 & -20.4\\
\cgpt  & \textbf{-12.0} & \textbf{-13.4} &  \textbf{-9.6} & \textbf{-12.8} & \textbf{-12.0}\\
\bottomrule
\end{tabular}

\textbf{dupe}\\
\begin{tabular}{lrrrrr}
\toprule
Model & \langp{en}{de} & \langp{en}{fr} & \langp{en}{ko} & \langp{en}{pt} & avg.\\
\midrule
\opus  & -39.5 & -45.9 & -26.5 & -54.5 & -41.6\\
\nllb  &  -9.3 &  -8.9 &  -3.6 & -14.7 & -9.1\\
\ti    &  -4.6 &  -5.0 &  -3.5 &  -5.2 & -4.6\\
\cgpt  &  \textbf{-3.7} &  \textbf{-4.0} &  \textbf{-3.2} &  \textbf{-3.1} & \textbf{-3.5}\\
\bottomrule
\end{tabular}

\textbf{key}\\
\begin{tabular}{lrrrrr}
\toprule
Model & \langp{en}{de} & \langp{en}{fr} & \langp{en}{ko} & \langp{en}{pt} & avg.\\
\midrule
\opus  & -48.8 & -62.9 & -29.0 & -64.6 & -51.3\\
\nllb  & -34.1 & -37.4 & -20.1 & -40.3 & -33.0\\
\ti    & -25.0 & -26.9 & -15.3 & -26.8 & -23.5\\
\cgpt  & \textbf{-11.8} & \textbf{-13.1} & \textbf{-10.5} & \textbf{-12.5} & \textbf{-12.0}\\
\bottomrule
\end{tabular}

    \caption{BLEU-slope on \flores{} for \langp{en}{xx}.}
    \label{table:flores-bleu-slope-en-xx}
\end{table}


\begin{table}[t]
    \small
    \centering

\textbf{swap}\\
\begin{tabular}{lrrrrr}
\toprule
Model & \langp{de}{en} & \langp{fr}{en} & \langp{ko}{en} & \langp{pt}{en} & avg.\\
\midrule
\opus  & -45.2 & -47.4 & -32.9 & -46.0 & -42.9\\
\nllb  & -25.3 & -26.6 & \textbf{-27.7} & -26.4 & -26.5\\
\ti    & -28.4 & -27.3 & -29.0 & -28.1 & -28.2\\
\cgpt  &  \textbf{-7.1} &  \textbf{-9.6} & -29.2 & \textbf{-10.9} & \textbf{-14.2}\\
\bottomrule
\end{tabular}

\textbf{drop}\\
\begin{tabular}{lrrrrr}
\toprule
Model & \langp{de}{en} & \langp{fr}{en} & \langp{ko}{en} & \langp{pt}{en} & avg.\\
\midrule
\opus  & -40.1 & -39.7 & -29.0 & -38.7 & -36.9\\
\nllb  & -23.8 & -23.6 & -26.6 & -24.1 & -24.5\\
\ti    & -22.3 & -20.9 & \textbf{-25.5} & -21.1 & -22.5\\
\cgpt  & \textbf{-10.0} &  \textbf{-9.3} & -26.0 & \textbf{-10.4} & \textbf{-13.9}\\
\bottomrule
\end{tabular}

\textbf{dupe}\\
\begin{tabular}{lrrrrr}
\toprule
Model & \langp{de}{en} & \langp{fr}{en} & \langp{ko}{en} & \langp{pt}{en} & avg.\\
\midrule
\opus  & -22.9 & -24.5 & -6.8 & -23.1 & -19.3\\
\nllb  &  -7.6 &  -8.6 &  -6.1 & -7.5 & -7.4\\
\ti    &  -5.1 &  -6.4 &  \textbf{-3.9} & -5.3 & -5.2\\
\cgpt  &  \textbf{-2.7} &  \textbf{-2.7} & -10.2 & \textbf{-2.8} & \textbf{-4.6}\\
\bottomrule
\end{tabular}

\textbf{key}\\
\begin{tabular}{lrrrrr}
\toprule
Model & \langp{de}{en} & \langp{fr}{en} & \langp{ko}{en} & \langp{pt}{en} & avg.\\
\midrule
\opus  & -43.6 & -45.5 & -34.5 & -46.0 & -42.4\\
\nllb  & -26.6 & -27.7 & -31.0 & -29.6 & -28.7\\
\ti    & -28.6 & -30.6 & -26.3 & -34.6 & -30.0\\
\cgpt  & \textbf{-12.1} & \textbf{-11.9} & \textbf{-22.6} & \textbf{-14.0} & \textbf{-15.2}\\
\bottomrule
\end{tabular}

    \caption{chrF-slope on \flores{} for \langp{xx}{en}.}
    \label{table:flores-chrf-slope-xx-en}
\end{table}

\begin{table}[t]
    \small
    \centering

\textbf{swap}\\
\begin{tabular}{lrrrrr}
\toprule
Model & \langp{en}{de} & \langp{en}{fr} & \langp{en}{ko} & \langp{en}{pt} & avg.\\
\midrule
\opus  & -50.0 & -56.8 & -38.0 & -55.7 & -50.1\\
\nllb  & -24.7 & -26.4 & -21.0 & -27.4 & -24.9\\
\ti    & -15.9 & -15.0 & -11.8 & -14.8 & -14.4\\
\cgpt  &  \textbf{-5.1} &  \textbf{-6.0} &  \textbf{-6.4} &  \textbf{-5.1} & \textbf{-5.7}\\
\bottomrule
\end{tabular}

\textbf{drop}\\
\begin{tabular}{lrrrrr}
\toprule
Model & \langp{en}{de} & \langp{en}{fr} & \langp{en}{ko} & \langp{en}{pt} & avg.\\
\midrule
\opus  & -48.3 & -53.3 & -36.6 & -53.5 & -47.9\\
\nllb  & -24.7 & -26.6 & -20.3 & -27.2 & -24.7\\
\ti    & -17.4 & -17.0 & -12.7 & -16.6 & -15.9\\
\cgpt  &  \textbf{-8.3} &  \textbf{-9.0} &  \textbf{-8.4} &  \textbf{-8.2} & \textbf{-8.5}\\
\bottomrule
\end{tabular}

\textbf{dupe}\\
\begin{tabular}{lrrrrr}
\toprule
Model & \langp{en}{de} & \langp{en}{fr} & \langp{en}{ko} & \langp{en}{pt} & avg.\\
\midrule
\opus  & -34.7 & -36.8 & -33.6 & -41.8 & -36.7\\
\nllb  &  -5.7 &  -5.8 &  -5.7 &  -8.6 & -6.4\\
\ti    &  -2.8 &  -2.6 &  -2.8 &  -2.7 & -2.7\\
\cgpt  &  \textbf{-1.9} &  \textbf{-2.1} &  \textbf{-2.5} &  \textbf{-1.3} & \textbf{-2.0}\\
\bottomrule
\end{tabular}

\textbf{key}\\
\begin{tabular}{lrrrrr}
\toprule
Model & \langp{en}{de} & \langp{en}{fr} & \langp{en}{ko} & \langp{en}{pt} & avg.\\
\midrule
\opus  & -50.0 & -56.6 & -37.9 & -55.9 & -50.1\\
\nllb  & -27.9 & -29.8 & -22.7 & -30.4 & -27.7\\
\ti    & -21.0 & -20.6 & -15.3 & -20.5 & -19.4\\
\cgpt  & \textbf{-8.4}  &  \textbf{-9.0} &  \textbf{-8.8} &  \textbf{-8.1} & \textbf{-8.6}\\
\bottomrule
\end{tabular}

    \caption{chrF-slope on \flores{} for \langp{en}{xx}.}
    \label{table:flores-chrf-slope-en-xx}
\end{table}

\end{document}